\documentclass[twoside,leqno,twocolumn]{article}  
\usepackage{ltexpprt}

\usepackage{url}

\usepackage{booktabs}
\usepackage{multirow}
\usepackage{tabularx}
\usepackage[table]{xcolor}
\newcolumntype{C}[1]{>{\raggedleft\let\newline\\\arraybackslash\hspace{0pt}}m{#1}}

\usepackage{multirow}
\usepackage{blindtext}
\usepackage{algorithm}
\usepackage{algorithmic}
\usepackage{multirow}
\usepackage{tabularx}
\usepackage[table]{xcolor}
\usepackage{blindtext}
\usepackage{graphicx}
\usepackage{verbatim}
\usepackage{amssymb}
\usepackage{color}

\usepackage{nameref}

\usepackage{amsmath}

\DeclareMathOperator*\argmin{\operatorname{argmin}}

\usepackage{times}
\usepackage{helvet}
\usepackage{courier}
\usepackage{latexsym}
\usepackage{amsmath}
\usepackage{algorithm}
\usepackage{algorithmic}
\usepackage{amssymb}
\usepackage{multirow}
\usepackage{tabularx}
\usepackage[table]{xcolor}
\usepackage{blindtext}
\usepackage{graphicx}
\usepackage{verbatim}
\usepackage{caption}
\usepackage{array}
\usepackage{url}
\usepackage{booktabs}
\usepackage{nomencl}
\usepackage{enumitem}
\setlist{nolistsep}

\newcommand\R{{\mathbb R}}

\newcommand\N{{\mathbb N}}

\usepackage[pdfborder={0 0 0}]{hyperref}

\begin{document}

\setlength{\abovedisplayskip}{0pt}
\setlength{\belowdisplayskip}{0pt}
\setlength{\abovedisplayshortskip}{0pt}
\setlength{\belowdisplayshortskip}{0pt}

\title{ Invariant Factorization of Time Series}
\author{Josif Grabocka \and Lars Schmidt-Thieme} 
\date{ \{ josif, schmidt-thieme \}@ismll.uni-hildesheim.de \\ 
ISMLL, University of Hildesheim, Germany 
}

\maketitle


\begin{abstract} \small\baselineskip=9pt 
Time-series classification is an important domain of machine learning and a plethora of methods have been developed for the task. In comparison to existing approaches, this study presents a novel method which decomposes a time-series dataset into latent patterns and membership weights of local segments to those patterns. The process is formalized as a constrained objective function and a tailored stochastic coordinate descent optimization is applied. The time-series are projected to a new feature representation consisting of the sums of the membership weights, which captures frequencies of local patterns. Features from various sliding window sizes are concatenated in order to encapsulate the interaction of patterns from different sizes. Finally, a large-scale experimental comparison against 6 state of the art baselines and 43 real life datasets is conducted. The proposed method outperforms all the baselines with statistically significant margins in terms of prediction accuracy.
\end{abstract}

\vspace{-0.1cm}
\section{Introduction}

Time-series classification is a pillar problem of machine learning and its existence spans over decades of research. Series data emerge in a myriad of application domains, from health-care and astronomy up to economics and botanics. In comparison to other types of data, time series exhibit a high degree of intra-class variability, where patterns occur shifted in time, distorted and scaled. Therefore traditionally strong classifiers, such as Support Vector Machines (SVM), fail to excel in terms of prediction accuracy \cite{Gudmundsson2008}. 

A series of attempts have been proposed to address the intra-class variations of time-series patterns. An early pioneer method called Dynamic Time Warping (DTW), (still considered competitive \cite{ding2008,Rakthanmanon2012}), computes the similarity among series by re-aligning the time indexes. The algorithm explores all the possible relative alignments of time indexes of two series and picks the one yielding the minimum overall distance~\cite{Keogh2000}. 

The research of time-series classification can be approximately categorized into distance metrics, invariant classifiers, feature extraction and bag-of-patterns streams. Distance metrics focus on defining measurements on the similarity of two series instances \cite{Keogh2000,Chen:2004:MLE:1316689.1316758,Chen2007,cid2013}. Invariant classifiers, on the other hand, aim at embedding similarities into classifiers. For instance, the invariant kernel functions have been applied to measure instance similarities in the projected space of a non-linear SVM \cite{Zhang2010,Gudmundsson2008}. Another paper proposes to generate all pattern variations as new instances and inflate the training set \cite{DBLP:conf/pkdd/GrabockaNS12}. The bag-of-patterns approach splits the time-series into local segments and collects statistics over the segments. Those local segments are converted into symbolic words and a histogram of the words' occurrences is built \cite{DBLP:journals/jiis/0001KL12,Lin:2009:FSS:1561638.1561679}. Another study constructs a supervised codebook generated from local patterns, which is used to create features for a random forest classifiers \cite{tsbf2013}.

In comparison to existing approaches this study proposes a new perspective. We assume that time series are generated by a set of latent (hidden) patterns which occur at different time stamps and different frequencies across instances. In addition those patterns might be convoluted and/or distorted to produce derived local patterns.

\begin{figure*}[!]
\centering
\includegraphics[scale=0.67, trim=2.0cm 13cm 2.0cm 6cm]{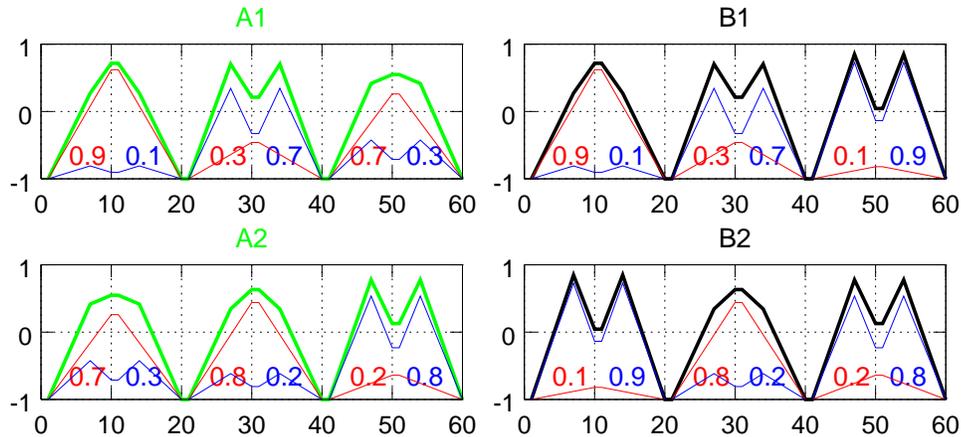}
\caption{Four series of two classes A=\{A1,A2\} and B=\{B1,B2\}, each generated as a convolution of latent patterns}
\label{syntheticDatasetFig}
\vspace{-0.3cm}
\end{figure*}

We would like to introduce the concept through the illustration of Figure~\ref{syntheticDatasetFig}. A synthetic dataset consists of two classes A (green) and B (black), each having two instances. All the time series are composed of three segments of 20 points, while each segment is a convolutional derivative of two latent patterns depicted in red and blue. In other words, each segment is a weighted sum of a single-peaked and double-peaked pattern. The shown coefficients of the convolution are degrees of membership that each local segment has to one of those two latent patterns.

Both Euclidean and DTW based nearest neighbor classifiers have 100\% error on a leave-one-out experiment on the dataset of Figure~\ref{syntheticDatasetFig}. As can be observed, instance A1 is closer to B1 than A2, and the same applies for all other series. In fact the rationale behind this dataset is that A has a higher frequency of the red single-peaked pattern, while B has a higher domination of the blue double-peaked pattern. The method presented in this paper detects the latent patterns, measures the degrees of membership and sums them up into a bag-of-pattern approach. Our approach converts the series of Figure~\ref{syntheticDatasetFig} into a new representation F, concretely: $F_{A1}=[{\color{red} 1.9}, {\color{blue} 1.1}], F_{A2}=[{\color{red} 1.7}, {\color{blue} 1.3}], F_{B1}=[{\color{red} 1.3}, {\color{blue} 1.7}], F_{B2}=[{\color{red} 1.1}, {\color{blue} 1.9}]$. A nearest neighbor classifier over the new representation F yields 0\% error.

In this paper, we will propose a method which detects a set of latent patterns for a time series dataset together with a convolutional degree of membership weights. The product of the membership weights with the patterns approximates the original segments. In contrast to the aforementioned synthetic example, real datasets have segments occurring at arbitrary locations and being of different sizes. Our method employs a sliding window approach to split the series into overlapping local segments and utilizes a factorization model to decompose the segments into latent patterns and weights. We formalize the objective function of the factorization and propose a stochastic coordinate descent technique in order to optimize the objective. The sum of the learned membership degrees is used to project the time series into a new representation. Ultimately, in order to resolve the scale invariance of the patterns, sums of memberships from different sliding window sizes are concatenated.

A throughout experimental comparison is conducted on 43 datasets of the UCR time-series collection against six state of the art baselines. Our method outperforms all the baselines with a statistically significant margin of improvement. Considering the evidence of our experimental results and our survey of related work, we conclude that our prediction accuracy figures are the best published in the realm of time-series classification, regarding the UCR collection of datasets.

\vspace{-0.1cm}

\section{Related Work}

Time-series classification has been elaborated in a vast number of occasions, therefore a complete survey of all the published papers is out of our scope. Instead, we will structure the related work into a set of categories and mention relevant prominent studies.

\vspace{-0.1cm}

\subsection{Distance Metrics and Invariant Classifiers:}

A significant portion of time-series research has centralized around the definition of accurate similarity metrics. The most popular of those approaches is the Dynamic Time Warping (DTW) measure \cite{Keogh2000}, which overcomes deficiencies of the $L_2$ norm distance by aligning the time indexes of two series instances. The similarity measure is typically plugged into a nearest neighbor classifier. DTW  produces competitive prediction accuracies \cite{ding2008,surveyRepresentations2013} and has been speed up using lower boundary heuristics \cite{Rakthanmanon2012}. 

Other similarity based distance metrics have extended the edit distance of strings into the time-series domain \cite{Chen:2004:MLE:1316689.1316758,Chen2007}. Furthermore, the longest common subsequence of time series has also been used as an indication of similarity \cite{Vlachos2002}. Moreover, similarities of sequential data have been measured using sparse spatial sample kernels \cite{kuksaSSSK2010}.  A state of the art method called complexity-invariant distance metric (CID) introduces the total variation regularization for time-series. CID significantly improves the accuracy of DTW \cite{DBLP:conf/sdm/BatistaWK11,cid2013}. 

Efforts have been dedicated on incorporating time-series variations into popular classifiers. For instance DTW has been used as a SVM kernel \cite{Gudmundsson2008}, even though the resulting kernel is not positive semi definite. Consecutively, another study has proposed a Gaussian elastic kernel \cite{Zhang2010}. A method which produces a semi-definite kernel is called global alignment kernels and builds an average statistics from all possible warping paths of time indexes \cite{ICML2011Cuturi}. In addition, another study has inflated the training set by adding new instances that represent variations of original training data \cite{DBLP:conf/pkdd/GrabockaNS12}.

\vspace{-0.2cm}
\subsection{Feature Extraction and Bag-of-patterns: }

Other researchers have emphasized the extraction of series features for boosting classification. Dimensionality reduction has been used to project the time series into a low-rank data space \cite{dimRedTimeSeries2001}, while a recent method incorporates class segregation into the projection \cite{DBLP:conf/aaai/GrabockaNS12}.

However, the most prominent state of the art technique for extracting time-series features is called shapelets mining. Shapelets represent the most discriminative series segment (or set of segments), which yields the maximal prediction accuracy \cite{rakthanmanon2013fast,DBLP:conf/kdd/MueenKY11}. A related study detects a set of shapelets and transforms the series data into a new representation, defined by the distance to those shapelets \cite{hills2013transform}.

A recent direction of research has drawn attention on the need to segment the time series into local patterns and measure the frequencies of patterns as classification features. For instance frequencies of time-series motifs have been fed into standard classifiers \cite{DBLP:conf/gfkl/BuzaS08}. Another attempt has focused on building histograms of local patterns represented as symbolic words \cite{Lin:2009:FSS:1561638.1561679}.  Those symbolic words are produced by a piecewise constant approximation technique called SAX \cite{Lin:2007:ESN:1285960.1285965}, while the frequencies of the SAX words are used ultimately for classification \cite{DBLP:journals/jiis/0001KL12,Lin:2009:FSS:1561638.1561679}. One similar bag-of-words approach has also been applied to long biomedical data \cite{Wang2013634}. Moreover, a bag-of-patterns study proposes to extract series segments of various lengths and positions and generate a supervised codebook of those patterns \cite{tsbf2013}. A random forest classifier has been trained over the extracted features. That study demonstrates considerable improvements over baselines in terms of prediction accuracy~\cite{tsbf2013}.

\vspace{-0.2cm}
\subsection{Factorization of Time Series}

There have been a few attempts in generating invariant time-series features through factorization. A shift-invariant sparse coding of signals has been proposed for reconstructing noisy or missing series segments \cite{LEW99}. In similar domains, sparse coding factorization has been applied for deriving shift and 2D rotation invariant features of hand writing data \cite{sparseCodingHandwriting2012}, and also invariant features of audio data \cite{conf/interspeech/HuangYHLH12}. Moreover, a temporal decomposition of multivariate streams has been used to discover patterns in patients' clinical events~\cite{Wang:2012:THT:2339530.2339605}. 


Our method differs from distance metrics principally. Instead of measuring the similarity of series, we project the data into a new representation, where similar instances are positioned close to each other. Furthermore, the proposed method distances away from existing bag-of-patterns methods because we learn a latent decomposition of patterns, instead of counting the occurrence of segments on the original time-series. Finally, our contributions over the existing factorization methods rely on (i) a novel approach in detecting both shift and scale invariant features for time series, and (ii) building a bag-of-patterns representation of the learned invariant features for a classification scenario.

\vspace{-0.2cm}
\section{Definitions and Notations}

\begin{enumerate}
\item{ \textbf{Time-series:} A time-series is an ordered sequence of point values. In a dataset of $N$ series instances, where each series has $Q$ points, we denote the series dataset as $T \in \R^{N \times Q}$.}
\item{\textbf{Sliding Window Segment:} A sliding window content of size $L \in \N$, is a series subsequence starting at a position $j \in \{1, \dots, Q-L\}$ of a series $i$ of dataset T, and is denoted as $S_{i,j}~\in~\R^{L}, \, S_{i,j}~:=~\left( T_{i,j},T_{i,j+1},\dots,T_{i,j+L-1}\right)$.}
\item{ \textbf{All Dataset Segments:}} The starting position of each sliding window segment is incremented by an offset $\delta=\{1,\dots,L\}$, therefore the maximum number of segments per series is defined as $M~:=~\frac{Q-L}{\delta}$. All the segments of a time-series datasets are denoted as $S \in \R^{N \times M \times L}$.
\item{ \textbf{Latent Patterns:} Our method mines for $K$-many latent patterns, each having the same size as one segment, i.e $L$. So, the latent patterns are denoted as $P \in \R^{K \times L}$.}
\item{ \textbf{Degrees of Membership:} Each instance of a dataset will be approximated via the product of latent patterns and the set of membership degrees to those patterns. Each segment of a series will have one membership weight to each of the $K$ latent patterns. Consequently, the degrees of membership of all time-series are defined as $D~\in~\R^{N \times M \times K}$.}
\end{enumerate}

\vspace{-0.2cm}
\section{Invariant Factorization of Time Series}
\subsection{Segmenting the Time-Series}

The series of the dataset are segmented in a sliding window approach having size $L$ and increment $\delta$. The segmentation of each series is described in Algorithm~\ref{segmentationAlg}. Once derived, the segments are normalized to mean 0 and deviation 1.

\begin{algorithm}[!]
\caption{SegmentSeries}
\begin{algorithmic}[1] 
\label{segmentationAlg} 
\REQUIRE $T \in \R^{ N \times Q}$, $L \in \N$, $\delta \in \N$
\ENSURE $S \in \R^{ N \times M \times L}$
\FOR{$ i=1,\dots,N, \; j=1,\dots,M$}
\FOR{$l=1,\dots,L$}
\STATE $S_{i,j,l} \leftarrow T_{i, \, \delta(j-1) + l -1}$
\ENDFOR
\STATE $S_{i,j} \leftarrow \mbox{normalize}(S_{i,j})$
\ENDFOR
\RETURN $S$
\end{algorithmic}
\end{algorithm}

\vspace{-0.3cm}

\subsection{Invariant Factorization Objective}

\begin{figure*}[!]
\centering
\includegraphics[scale=0.73, trim=2.0cm 9.5cm 2.0cm 6cm]{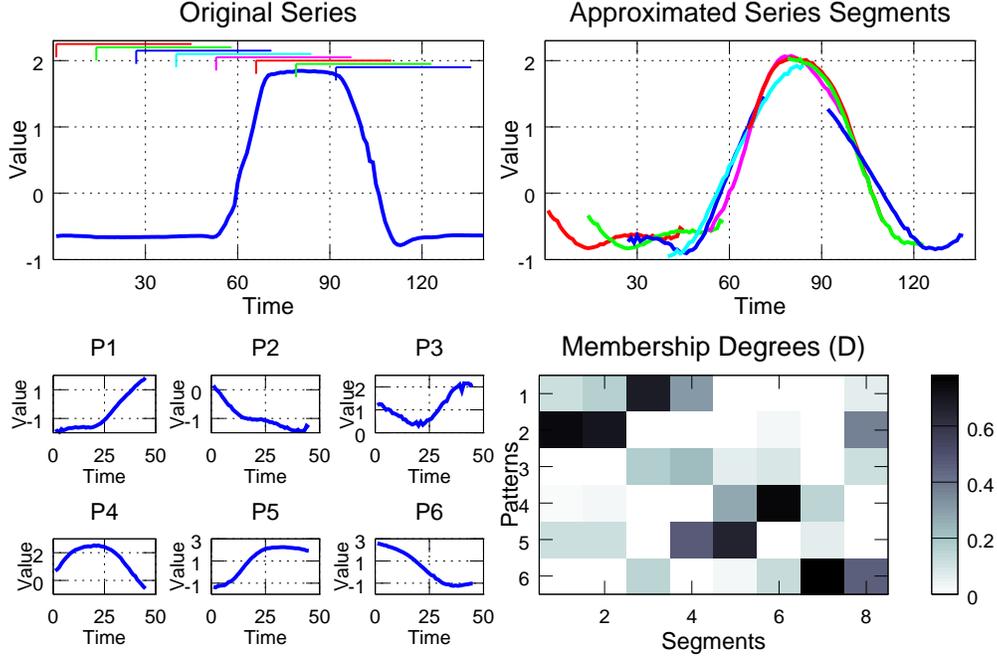}
\caption{A factorized instance of the Gun Point dataset with parameters $K=6,L=45,\delta=13,\lambda_P=1$}
\label{mechanismFig}
\vspace{-0.3cm}
\end{figure*}

\vspace{-0.2cm}

The objective of the invariant factorization relies on approximating every segment of a series segment as a product of the defined latent patterns $P$ and the membership degrees $D$. The objective function is described in Equation~\ref{infaLoss}. 

\begin{align}
\label{infaLoss}
\argmin_{D,P} &\sum_{i=1}^N \sum_{j=1}^M \sum_{l=1}^L \left(S_{i,j,l} - \sum_{k=1}^K D_{i,j,k} P_{k,l}\right)^2 \\ \nonumber
& + \lambda_P \sum_{k=1}^K \sum_{l=1}^L  P_{k,l}^2  \\ \nonumber
\mbox{{Subject To:} }& \\ \nonumber
& \sum_{k=1}^K D_{i,j,k} = 1, \;\;\; D_{i,j,k} \ge 0, \;\;\; \forall i,j,k
\end{align}
\vspace{0.1cm}

The objective function is composed of two loss terms and one constraint. Firstly, the latent patterns $P$ and the memberships $D$ should approximate the normalized segments of the series dataset. Therefore, minimizing the L2 norm of the reconstruction error achieves the goal. In addition, a second regularization loss term is added in order to prohibit the patterns $P$ from over-fitting. A hyper-parameter $\lambda_P$ controls the degree of regularization. Finally, we impose equality and positivity constraints on the membership degrees. The membership degrees of every segment $D_{i,j}$ sum-up to one, because each segment needs to have the same impact factor. Otherwise, in a bag-of-patterns representation of series, different segments would have different scales of memberships. The positivity constraint, on the other hand, prohibits non-interpretable negative memberships.

We would like to illustrate the invariant factorization objective with a concrete illustration, shown in Figure \ref{mechanismFig}. A learned decomposition, as in Equation~\ref{infaLoss}, is depicted for the Gun Point dataset. On the left top, a series instance is presented, while the dataset's latent patterns and the membership degrees of the instance are found below. The product of the patterns and memberships yield the series approximation shown in the right top chart. The series is split into 8 overlapping segments of size 45, each starting at an offset of 13 points. For instance, the 7-th segment starts at 79 and has a high membership value to the 6-th pattern, which matches the descending structure. However, please note that other patterns also contribute with smaller membership degrees (patterns 4 and 5) in order to fit exactly the original segment content.

\vspace{-0.2cm}
\subsection{Learning the Patterns and Memberships}

In order to learn the latent patterns and the memberships we are going to optimize the objective function of Equation~\ref{infaLoss} via \emph{\bf stochastic coordinate descent}, which operates by updating each cell of $D,P$ in the direction of the first derivative of the objective. 

\vspace{-0.1cm}
\subsubsection{Update Rules for Latent Patterns}

In order to compute the update rules for the patterns, we first define the error in approximating a point $l$ of the segment $j$, in time-series $i$, as $\xi_{i,j,l}$. 

\vspace{-0.3cm}

\begin{align}
\nonumber
&\mbox{Let } \xi_{i,j,l} := S_{i,j,l} - \sum_{k=1}^K D_{i,j,k} P_{k,l} \\ \nonumber
&P_{k,l}^* := \argmin_{z} \lambda_P z^2 + \sum_{i,j} \left( \xi_{i,j,l} + D_{i,j,k} P_{k,l} - D_{i,j,k} z \right)^2 \\ \nonumber
&2\lambda_P P_{k,l}^*  - 2\sum_{i,j} \left(\xi_{i,j,l} + D_{i,j,k} ( P_{k,l} - P_{k,l}^*)\right) D_{i,j,k} = 0
\end{align}

\vspace{-0.0cm}

Subsequently the optimal value of every point $l$ of a latent pattern $k$ is denoted as $P_{k,l}^*$ and is found by solving the first derivative in a coordinate descent way. Therefore the optimal value of $P_{k,l}$ is defined in Equation~\ref{udateRulePattern}. Please note that the error values don't have to be recomputed for each point over all latent patterns, instead we can incrementally update the error terms. Equation~\ref{udateRulePatternError} refreshes the error terms after the change of the pattern value.

\vspace{-0.1cm}

\begin{align}
\label{udateRulePattern}
& P_{k,l}^* := \frac{ \sum_{i,j} \left( \xi_{i,j,l} + D_{i,j,k} P_{k,l} \right) D_{i,j,k} }{\lambda_P + \sum_{i,j}D_{i,j,k}^2} \\ 
\label{udateRulePatternError}
& \xi_{i,j,l} \leftarrow \xi_{i,j,l} - (P_{k,l}^* - P_{k,l}) D_{i,j,k}
\end{align}

\subsubsection{Update Rules for Membership Degrees}

The update rules for the membership degrees needs to preserve an equality constraint, which enforce the memberships of a segment to sum to one. Therefore any direct update of a membership $D_{i,j,k}$ will violate the constraint. In order to avoid this bottleneck, we propose to update the memberships in pairs, inspired by a similar strategy known as the Sequential Minimal Optimization algorithm \cite{Platt:1999:FTS:299094.299105}. The idea is to draw two random membership weights $D_{i,j,k},D_{i,j,w}$ and update them such that their sum, denoted $Q=D_{i,j,k}+D_{i,j,w}$, remains equal before and after the updates. In that way, if we increase one membership, then the other would have to decrease and vice versa, while the aim is to find the combination which yield the smallest approximation error. Therefore, the optimal value of $D_{i,j,k}$, will be denoted by $D_{i,j,k}^*$ and is algebraically derived as the solution of the first derivative of our objective function.

\vspace{-0.1cm}
\begin{align}
\nonumber
&D_{i,j,k}^* = \argmin_{z} \sum_l ( \xi_{i,j,l} + D_{i,j,k} P_{k,l} + D_{i,j,w} P_{w,l} - \\ \nonumber
&\;\;\;\;\;\;\;\;\;\;\;\;\;\;\;\;\;\;\;\;\;\;\;\;\;\;\;\;\;\;\;\; Q P_{w,l} + z ( P_{w,l} - P_{k,l} ))^2 \\ \nonumber
&D_{i,j,k}^* = \frac{- \sum_l \left(\xi_{i,j,l} - D_{i,j,k} \left( P_{w,l} - P_{k,l} \right) \right) \left( P_{w,l} - P_{k,l} \right) }{\sum_l \left( P_{w,l} - P_{k,l} \right)^2} 
\end{align}
\vspace{0.1cm}

Once the optimal value $D_{i,j,k}^*$ is defined then we have to ensure the constraints. Equation~\ref{updateRuleDijk} crops the optimal value to be nonnegative and not exceed the sum of the membership pairs. The error term is refreshed to include the changes of both memberships of the pair in Equation~\ref{updateRuleEijl}. As a last step we can commit the optimal values, by preserving their sum before the updates. As Equation~\ref{updateRuleCommit} shows, the best value of $D_{i,j,w}$ can be deduced from the optimal value of $D_{i,j,k}$.

\vspace{-0.1cm}

\begin{eqnarray}
\label{updateRuleDijk}
&& D_{i,j,k}^* \leftarrow \mbox{max}(0, \mbox{min}(Q, D_{i,j,k}^*)) \\
\label{updateRuleEijl}
&& \xi_{i,j,l} \leftarrow \xi_{i,j,l} - (D_{i,j,k}^* - D_{i,j,k}) P_{k,l}, \mbox{  and }   \\ \nonumber 
&& \xi_{i,j,l} \leftarrow \xi_{i,j,l} - (Q - D_{i,j,k}^* - D_{i,j,w}) P_{w,l}  \\ 
\label{updateRuleCommit}
&&  D_{i,j,k} \leftarrow D_{i,j,k}^*, \;\; D_{i,j,w} \leftarrow Q - D_{i,j,k}^* 
\end{eqnarray}

\subsection{Efficient Initialization}

Since the objective function of Equation~\ref{infaLoss} is nonlinear in terms of $P$ and $D$ together, then a coordinate descent optimization is not guaranteed to avoid local optima. Therefore, good initial values of the patterns and the memberships are crucial for the learning process. The intuition leads into assigning some of the  segments as initial patterns, however it is not obvious which of them provide the best initialization. 

The answer is addressed via a technique utilized to find the initial centroids in a clustering setup \cite{Arthur:2007:KAC:1283383.1283494}. The patterns (analogy to centroids) are initialized to segments with a probability proportional to the distance to all the other segments \cite{Arthur:2007:KAC:1283383.1283494}. Therefore, we are assured to pick centroid segments which are evenly distributed across the space of all series segments. The initialization steps are detailed in Algorithm~\ref{initializationAlg}. Please note that the first pattern has to be drawn randomly in a uniform distribution, while the other patterns are chosen randomly from the dataset segments based on the probability of their distance to the existing patterns. The function $\mathcal{C}$ measures the distance of a segment to the closest existing pattern. 

\vspace{-0.1cm}

\begin{algorithm}[h]
\caption{Initialize}
\begin{algorithmic}[1] 
\label{initializationAlg}
\REQUIRE $S \in \R^{N\times M \times L}, L \in \N, K \in \N$
\ENSURE $D \in \R^{N \times M \times K }, P \in \R^{K \times L}$
\STATE $P_{1} \leftarrow S_{i',j'}, \mbox{ drawn } i',j' \sim \mathcal{U}(N,M)$ 
\FOR{$ k=2,\dots,K$}
\STATE $P_{k} \leftarrow S_{i',j'}, \mbox { with probability weights } \frac{ \mathcal{C} \left(S_{i',j'}\right)^2}{\sum_{i,j} \mathcal{C} \left(S_{i,j}\right) ^2} $
\ENDFOR
\FOR{$ i=1,\dots,N; \; j=1,\dots,M$}
\STATE $k' = \argmin_{k \in \{1,\dots,K\}} || S_{i,j}- P_k ||^2$
\STATE $D_{i,j,k} \leftarrow \begin{cases} 1 & k = k' \\ 
0 & k \ne k' \end{cases}, \;\; k=1,\dots,K$
\ENDFOR
\RETURN $D, P$
\end{algorithmic}
\end{algorithm}

\vspace{-0.1cm}

The initialization of the membership degrees is more trivial than patterns. The degree index $k'$ denotes that pattern $P_{k'}$ is the closest to segment $S_{i,j}$ and its membership $D_{i,j,k'}$ is set to 1, while all the other membership degrees are initialized to zero.

\subsection{Learning Algorithm}

\begin{algorithm}[!t]
\caption{InvariantFactorization}
\begin{algorithmic}[1] 
\label{invariantFactorizationAlg}
\REQUIRE $T \in \R^{ N \times Q}, L \in \N, \delta \in \N, K \in \N, \lambda_P \in \R, \mathcal{I}\in \N$
\ENSURE $D \in \R^{N \times M \times K }, P \in \R^{K \times L}$
\STATE $S \leftarrow $ SegmentSeries$(T,L,\sigma)$
\STATE $(D,P) \leftarrow $ Initialize$(S,L,K)$
\STATE \COMMENT{Initialize the errors}
\FOR{$ \forall i \in \N_1^N,  \forall j \in \N_1^M,  \forall l \in \N_1^L$}
\STATE $\xi_{i,j,l} := S_{i,j,l} - \sum_{k=1}^K D_{i,j,k} P_{k,l}$
\ENDFOR
\STATE \COMMENT{Update the patterns\&memberships iteratively}
\FOR{ iteration = $1,\dots,\mathcal{I}$ }
\STATE \COMMENT{Update all degrees of membership}
\FOR{$ \forall i \in \N_1^N,  \forall j \in \N_1^M \mbox{ randomly }$}
\FOR{$1,\dots,K$, \COMMENT{Draw K-many pairs} }
\STATE $k,w \sim \mathcal{U}(K,K), \mbox{ s.t. } D_{i,j,k} + D_{i,j,w} \ne 0$
\STATE $Q \leftarrow D_{i,j,k} + D_{i,j,w} $
\STATE \COMMENT{Solve and crop the optimal memberships}
\STATE $D_{i,j,k}^* = \frac{- \sum_l \left(\xi_{i,j,l} - D_{i,j,k} \left( P_{w,l} - P_{k,l} \right) \right) \left( P_{w,l} - P_{k,l} \right) }{\sum_l \left( P_{w,l} - P_{k,l} \right)^2} $
\STATE $D_{i,j,k}^* \leftarrow \mbox{max}\left(0, \mbox{min}(Q, D_{i,j,k}^*)\right)$
\STATE \COMMENT{Update the error terms}
\FOR{$l=1,\dots,L $}
\STATE $\xi_{i,j,l} \leftarrow \xi_{i,j,l} - (D_{i,j,k}^* - D_{i,j,k}) P_{k,l}$
\STATE $\xi_{i,j,l} \leftarrow \xi_{i,j,l} - (Q - D_{i,j,k}^* - D_{i,j,w}) P_{w,l} $
\ENDFOR
\STATE \COMMENT{Commit the values of the pair}
\STATE $D_{i,j,k} \leftarrow D_{i,j,k}^*$
\STATE $D_{i,j,w} \leftarrow Q - D_{i,j,k}^*$
\ENDFOR
\ENDFOR
\STATE \COMMENT{Update all patterns}
\FOR{$ \forall k \in \N_1^K; \; \forall l \in \N_1^L, \mbox{ randomly } $}
\STATE $P_{k,l}^* = \frac{ \sum_{i,j} \left( \xi_{i,j,l} + D_{i,j,k} P_{k,l} \right) D_{i,j,k} }{\lambda_P + \sum_{i,j}D_{i,j,k}^2}$
\STATE \COMMENT{Update the error terms}
\FOR{$i=1,\dots,N; \; j=1,\dots,M$}
\STATE $\xi_{i,j,l} \leftarrow \xi_{i,j,l} - (P_{k,l}^* - P_{k,l}) D_{i,j,k}$
\ENDFOR
\STATE \COMMENT{Commit the pattern's point value}
\STATE $P_{k,l} \leftarrow P_{k,l}^*$
\ENDFOR
\ENDFOR
\RETURN $D, P$
\end{algorithmic}
\end{algorithm}
\vspace{-0.2cm}

Algorithm~\ref{invariantFactorizationAlg} finally combines all the steps of the factorization process. In the beginning, the memberships and the patterns are initialized using Algorithm~\ref{initializationAlg}. Next the errors are initialized, then the coordinate descent technique updates all the parameters in a number of iterations, denoted as a hyper-parameter $\mathcal{I}$. Subsequently, the degrees of membership and the patterns are learned by setting the aforementioned optimal values. The membership and pattern indexes are visited in random order to speed up the convergence.

\vspace{-0.2cm}
\subsection{A New Invariant Representation}

The final representation will sum the membership degrees in a bag-of-patterns strategy. It enables a quantification of which local patterns appear in a series and how often. The shift invariance is achieved by segmenting the series in a sliding window approach and the scale invariance is addressed using different sliding window sizes. Algorithm~\ref{shiftScaleInvariantRepresentationAlg} describes the algorithmic steps. The algorithm iterates over $\Phi$ many different scales of an initial sliding windows size $L$ and solves an invariant factorization from Algorithm~\ref{invariantFactorizationAlg} per each size. The frequencies of the learned memberships are summed up for all $K$ patterns and the procedure is repeated for every sliding window size. Finally each time series contains $K \Phi$ many features, which denote the frequencies of patterns at different sizes and positions.

The new representation will be used for classification, instead of the original time series. We deployed a polynomial kernel Support Vector Machines, because we need to capture the interaction among features, i.e. the interaction among patterns of various sizes.


\begin{algorithm}[h]
\caption{InvariantRepresentation}
\begin{algorithmic}[1] 
\label{shiftScaleInvariantRepresentationAlg}
\REQUIRE $T \in \R^{ N \times Q}, L \in \N, \delta \in \N, K \in \N, \lambda_P \in \R, \mathcal{I} \in \N, \Phi \in N$
\ENSURE $F \in \R^{N \times (K \Phi) }$
\FOR{$ s = 1,\dots,\Phi $}
\STATE $L' \leftarrow L \cdot s$ 
\STATE $D \leftarrow$ InvariantFactorization$(T,\underline{L'}, \delta,K,\lambda_P,I)$ 
\FOR{$i = 1,\dots,N; \; k=1,\dots,K$}
\STATE $M \leftarrow \frac{Q-L'}{\delta}$
\STATE $F_{i, k + (s-1)K} \leftarrow \sum_{j=1}^M D_{i,j,k}$ 
\ENDFOR
\ENDFOR
\RETURN $F$
\end{algorithmic}
\end{algorithm}

\vspace{-0.2cm}

\subsection{Algorithmic Complexity} The run-time complexity of the method is dominated by the updates of memberships and has an order $O(N M K L \mathcal{I})$. Concretely our method needs 48.4 hours to compute on the StarLightCurves (the largest) dataset, while for instance DTW needs 87 hours. The space complexity of our method depends on the storage of the segments $S$ and the memberships $D$, which is $O(NM \max(K,L))$.

\vspace{-0.1cm}

\section{Experimental Results}
\label{experimentsLabel}

\subsection{Baselines} We compared the prediction accuracy of our method, denoted Invariant Factorization (\textbf{INFA}), against the following six state of the art baselines:

\begin{itemize}
\item{\bf TSBF:} The bag-of-features framework for time series (TSBF) uses a supervised codebook to extract features for a random forest classifier~\cite{tsbf2013}.
\item{\bf SSSK:} Sparse Spatial Similarity Kernel (SSSK) measures sequence similarity through sampling sequence features at different resolutions~\cite{kuksaSSSK2010}.
\item{\bf BOW:} The Bag of Words (BOW) method decomposes the series into local SAX words and uses a histogram representation of words as the new feature representation \cite{DBLP:journals/jiis/0001KL12,Lin:2009:FSS:1561638.1561679}.
\item{\bf DTW:} Dynamic Time Warping (DTW) computes the best alignment of time indexes resulting in the mininal distance \cite{Keogh2000, Rakthanmanon2012}.
\item{\bf CID:} The complexity invariant distance (CID) adds a L2-based total variation regularization term into the DTW distance \cite{cid2013}.
\item{\bf FSH:} Fast shapelet (FSH) extracts the most discriminative segment of the series dataset, such that the distance from the dataset instances to the optimal shapelet can be used as a feature for classification~\cite{rakthanmanon2013fast}.
\end{itemize}

\begin{table*}[!]
  \centering
  \caption{Error Rates - Comparison of Prediction Accuracies on the UCR Collection of Datasets}
    \begin{tabular}{|c||r|r|r|r||c|c|c|c|c|c|c|}
    \hline
    \multicolumn{1}{|l||}{\textbf{Dataset}} & \multicolumn{1}{l|}{\textbf{Cls.}} & \multicolumn{1}{c|}{\textbf{Train}} & \multicolumn{1}{c|}{\textbf{Test}} & \multicolumn{1}{c||}{\textbf{Len.}} & \multicolumn{1}{c|}{\textbf{INFA}} & \textbf{TSBF} & \textbf{SSSK} & \textbf{BOW} & \textbf{DTW} & \textbf{CID} & \textbf{FSH} \\ \hline \hline
    50words & 50    & 450   & 455   & 270   & 0.215 & \textbf{0.209} & 0.488 & 0.316 & 0.310 & 0.226 & 0.557 \\ \hline
    Adiac & 37    & 390   & 391   & 176   & 0.315 & \textbf{0.245} & 0.575 & 0.325 & 0.396 & 0.379 & 0.514 \\ \hline
    Beef  & 5     & 30    & 30    & 470   & 0.333 & 0.287 & 0.633 & \textbf{0.267} & 0.500 & 0.467 & 0.447 \\ \hline
    CBF   & 3     & 30    & 900   & 128   & \textbf{0.000} & 0.009 & 0.090 & 0.048 & 0.003 & 0.001 & 0.053 \\ \hline
    Chlorine & 3     & 467   & 3840  & 166   & 0.451 & \textbf{0.336} & 0.428 & 0.405 & 0.352 & 0.351 & 0.417 \\ \hline
    CinC\_ECG & 4     & 40    & 1380  & 1639  & 0.159 & 0.262 & 0.438 & 0.164 & 0.349 & \textbf{0.054} & 0.174 \\ \hline
    Coffee & 2     & 28    & 28    & 286   & \textbf{0.000} & 0.004 & 0.071 & 0.036 & 0.179 & 0.179 & 0.068 \\ \hline
    Cricket\_X & 12    & 390   & 390   & 300   & \textbf{0.197} & 0.278 & 0.585 & 0.305 & 0.223 & 0.249 & 0.527 \\ \hline
    Cricket\_Y & 12    & 390   & 390   & 300   & 0.208 & 0.259 & 0.654 & 0.313 & 0.208 & \textbf{0.197} & 0.505 \\ \hline
    Cricket\_Z & 12    & 390   & 390   & 300   & \textbf{0.200} & 0.263 & 0.574 & 0.295 & 0.208 & 0.205 & 0.547 \\ \hline
    Diatom & 4     & 16    & 306   & 345   & \textbf{0.007} & 0.126 & 0.173 & 0.111 & 0.033 & 0.065 & 0.117 \\ \hline
    ECG200 & 2     & 100   & 100   & 96    & 0.140 & 0.145 & 0.220 & \textbf{0.110} & 0.230 & \textbf{0.110} & 0.227 \\ \hline
    ECGFiveDays & 2     & 23    & 861   & 136   & \textbf{0.000} & 0.183 & 0.360 & 0.164 & 0.232 & 0.218 & 0.004 \\ \hline
    FaceAll & 14    & 560   & 1690  & 131   & 0.237 & 0.234 & 0.369 & 0.238 & 0.192 & \textbf{0.144} & 0.411 \\ \hline
    FaceFour & 4     & 24    & 88    & 350   & \textbf{0.000} & 0.051 & 0.102 & 0.102 & 0.170 & 0.125 & 0.090 \\ \hline
    FacesUCR & 14    & 200   & 2050  & 131   & \textbf{0.083} & 0.090 & 0.356 & 0.137 & 0.095 & 0.102 & 0.328 \\ \hline
    Fish  & 7     & 175   & 175   & 463   & \textbf{0.063} & 0.080 & 0.177 & 0.029 & 0.167 & 0.154 & 0.197 \\ \hline
    Gun\_Point & 2     & 50    & 150   & 150   & \textbf{0.007} & 0.011 & 0.133 & 0.407 & 0.093 & 0.073 & 0.061 \\ \hline
    Haptics & 5     & 155   & 308   & 1092  & 0.536 & \textbf{0.488} & 0.591 & 0.630 & 0.623 & 0.571 & 0.616 \\ \hline
    InlineSkate & 7     & 100   & 550   & 1882  & 0.622 & 0.603 & 0.729 & 0.629 & 0.616 & \textbf{0.586} & 0.741 \\ \hline
    ItalyPower & 2     & 67    & 1029  & 24    & 0.049 & 0.096 & 0.101 & \textbf{0.044} & 0.050 & \textbf{0.044} & 0.095 \\ \hline
    Lighting2 & 2     & 60    & 61    & 637   & 0.148 & 0.257 & 0.393 & 0.328 & \textbf{0.131} & \textbf{0.131} & 0.295 \\ \hline
    Lighting7 & 7     & 70    & 73    & 319   & \textbf{0.233} & 0.262 & 0.438 & 0.370 & 0.274 & 0.260 & 0.403 \\ \hline
    MALLAT & 8     & 55    & 2345  & 1024  & \textbf{0.028} & 0.037 & 0.153 & 0.098 & 0.066 & 0.075 & 0.033 \\ \hline
    MedicalImages & 10    & 381   & 760   & 99    & 0.275 & 0.269 & 0.463 & 0.401 & 0.263 & \textbf{0.258} & 0.433 \\ \hline
    MoteStrain & 2     & 20    & 1252  & 84    & \textbf{0.097} & 0.135 & 0.166 & 0.177 & 0.165 & 0.205 & 0.217 \\ \hline
    OliveOil & 4     & 30    & 30    & 570   & 0.367 & \textbf{0.090} & 0.300 & 0.233 & 0.133 & 0.167 & 0.213 \\ \hline
    OSULeaf & 6     & 200   & 242   & 427   & 0.190 & 0.329 & 0.326 & \textbf{0.153} & 0.409 & 0.372 & 0.359 \\ \hline
    Sony  & 2     & 20    & 601   & 70    & \textbf{0.163} & 0.175 & 0.376 & 0.409 & 0.275 & 0.185 & 0.315 \\ \hline
    SonyII & 2     & 27    & 953   & 65    & \textbf{0.075} & 0.196 & 0.339 & 0.154 & 0.169 & 0.123 & 0.215 \\ \hline
    StarLightCurves & 3     & 1000  & 8236  & 1024  & 0.023 & 0.022 & 0.135 & \textbf{0.021} & 0.093 & 0.066 & 0.063 \\ \hline
    SwedishLeaf & 15    & 500   & 625   & 128   & 0.078 & \textbf{0.075} & 0.339 & 0.125 & 0.210 & 0.117 & 0.269 \\ \hline
    Symbols & 6     & 25    & 995   & 398   & 0.036 & \textbf{0.034} & 0.184 & 0.088 & 0.050 & 0.059 & 0.068 \\ \hline
    synthetic\_control & 6     & 300   & 300   & 60    & 0.010 & 0.008 & 0.067 & 0.017 & \textbf{0.007} & 0.027 & 0.081 \\ \hline
    Trace & 4     & 100   & 100   & 275   & \textbf{0.000} & 0.020 & 0.300 & \textbf{0.000} & \textbf{0.000} & 0.010 & 0.002 \\ \hline
    Two\_Patterns & 4     & 1000  & 4000  & 128   & 0.005 & 0.001 & 0.087 & 0.010 & \textbf{0.000} & 0.004 & 0.114 \\ \hline
    TwoLeadECG & 2     & 23    & 1139  & 82    & \textbf{0.005} & 0.046 & 0.257 & 0.248 & 0.096 & 0.138 & 0.090 \\ \hline
    uWaveX & 8     & 896   & 3582  & 315   & 0.177 & \textbf{0.164} & 0.358 & 0.242 & 0.273 & 0.211 & 0.293 \\ \hline
    uWaveY & 8     & 896   & 3582  & 315   & \textbf{0.225} & 0.249 & 0.493 & 0.352 & 0.366 & 0.278 & 0.392 \\ \hline
    uWaveZ & 8     & 896   & 3582  & 315   & 0.229 & \textbf{0.217} & 0.439 & 0.325 & 0.342 & 0.293 & 0.364 \\ \hline
    Wafer & 2     & 1000  & 6174  & 152   & \textbf{0.003} & 0.004 & 0.029 & 0.010 & 0.020 & 0.006 & 0.004 \\ \hline
    WordsSynonyms & 25    & 267   & 638   & 270   & 0.303 & 0.302 & 0.553 & 0.371 & 0.351 & \textbf{0.243} & 0.594 \\ \hline
    yoga  & 2     & 300   & 3000  & 426   & \textbf{0.112} & 0.149 & 0.172 & 0.145 & 0.164 & 0.156 & 0.269 \\ \hline
    \multicolumn{5}{|l||}{\bf Absolute Wins}   & \textbf{19.33} & \textbf{9.00} & \textbf{0.00} & \textbf{4.33} & \textbf{2.83} & \textbf{7.5} & \textbf{0} \\ \hline \hline
    \multicolumn{6}{|l|}{\bf INFA One-to-one Wins}      & \textbf{26}    & \textbf{41}    & \textbf{34}    & \textbf{34}    & \textbf{31}  & \bf 41 \\ \hline
    \multicolumn{6}{|l|}{\bf INFA One-to-one Draws}     & \bf 0     & \bf 0     & \bf 1     & \bf 1     & \bf 0  & \bf 0 \\ \hline
    \multicolumn{6}{|l|}{\bf INFA One-to-one Losses}    & \bf 17    & \bf 2     & \bf 8    & \bf 8    & \bf 12  & \bf 2 \\
    \hline \hline
    \multicolumn{6}{|l|}{\bf Wilcoxon Test (p values) (Stat. significant for $p \le 0.05$)}    & \bf 0.028    & \bf 0.000     & \bf 0.000    & \bf 0.000    & \bf 0.005  & \bf 0.000 \\
    \hline
    \end{tabular}%
  \label{resultsTable}%
\end{table*}%

\vspace{-0.2cm}
\subsection{Setup and Reproducibility}
We conducted a large-scale experimentation in 43 time-series dataset from the UCR collection\footnote{\url{www.cs.ucr.edu/~eamonn/time_series_data}}. Our protocol complied to the default train/test split of the data, which is an established benchmark split and is used by the baselines. The metric of comparison is the error rate, i.e. the misclassification rate. Table~\ref{resultsTable} shows the datasets used for experimentation together with the number of classes, the number of training instances, the number of testing instances and the length of the series.

Our method has a relatively high number of hyper-parameters, however most of them can be analytically adjusted. Since all the segments are normalized, then the latent patterns should also have mean 0 and standard deviation 1. Therefore, $\lambda_P=1$ by the definition of the Tikhonov regularization. We searched for four different sliding windows sizes, i.e. $\Phi=4$ and $L=20\% \mbox{ of } Q$, so $ L' \in \{20\%, 40\%, 60\%, 80\% \} \mbox{ of } Q$. The number of latent patterns needs to be set large enough to avoid underfitting and was set to $K=50\% \mbox{ of } Q$. A fine grained sliding window offset was applied as $\delta=5\% \mbox{ of } L$. However, in order to ensure the scalability for the four largest datasets (Cin\_ECG, InlineSkate, MALLAT, StarLightCurves) we set their parameters to $K=10\% \mbox{ of } Q$ and $ \delta=20\% \mbox{ of } L$. The maximum number of iterations was set to $\mathcal{I}=15$.  The applied classifier was a polynomial kernel SVM with a polynomial degree being 3 and the complexity parameter 1, which are competitive SVM settings for the UCR collection \cite{DBLP:conf/pkdd/GrabockaNS12}. Since the algorithm is based on a probabilistic initialization, it might be possible that it converges to different closeby optima in each execution. However, in our experiments, those optima were very close and the final prediction accuracy results have insignificant differences. {\bf The authors are devoted to promote full reproducibility, therefore the source code, the data and instructions are publicly available\footnote{\url{http://fs.ismll.de/publicspace/InvariantFactorization/}}. }

\subsection{Results}

The error rate results of the six state of the art baselines and our method INFA are presented in Table~\ref{resultsTable}. The best performing method for each dataset (row) is emphasized in bold. In order to compare multiple classifiers across a large number of datasets we follow the established benchmarks of counting wins and \emph{Wilcoxon's Signed-Rank} test for statistical significance \cite{Demsar:2006:SCC:1248547.1248548}. To be fair with the baselines, we retrieved the results from the baselines' publications \cite{tsbf2013,cid2013,rakthanmanon2013fast} over the {\bf same} data splits as INFA. In addition, we verified the published results of the baselines with our own experimental checkups.

Three comparative figures are conducted, the first of which counts the absolute number of wins. Each dataset awards a total value of 1, which is split into equal fractions in case methods have equal error rate scores. The "Absolute wins" row, in the bottom of the table, counts the datasets where a method has the best prediction accuracy. As can be trivially deduced, our method has a clear superiority in terms of absolute wins, scoring 19.37 wins against 9.00 wins of the second best method. In addition INFA outperforms by large margins all the baselines in an one-to-one comparisons of wins. INFA has more wins, yet the predominant analysis is whether or not those wins represent statistically significant differences. Each cell on the bottom row represents the p value of the Wilcoxon Signed-Rank test on the error rate values of INFA against each baseline. Our method has a statistically significant difference over the error results of all baselines with a two-tailed hypothesis and the standard significance level of $95 \%$ confidence ($p \le 0.05$).

Based on our survey of related work, the results presented in this study are the best published prediction accuracy scores in the realm of time-series classification, with respect to the UCR collection of datasets.

\vspace{-0.1cm}
\section{Conclusions}
In this study we presented Invariant Factorization, a method that initially decomposes the time series into a set of overlapping segments via a sliding window approach. The segments are approximated by learning a set of latent patterns and degrees of memberships of each segment to each pattern. We formalized the factorization as a constraint objective function and proposed a stochastic coordinate descent method to solve it. The new representation of time series are the sums of the membership weights, which represent frequencies of local patterns. Features from various sliding window sizes were concatenated to encapsulate interaction among patterns of various scales. Finally we conducted a thorough experimental comparison against 6 state of the art baselines in 43 real-life time series datasets. Our method outperforms all the baselines with statistically significant margins and marks the best published results in the realm of time-series classification, regarding the UCR collection of datasets.


%

\vspace{-0.3cm}

\bibliography{allrelated}

\begin{thebibliography}{10}

\bibitem{Arthur:2007:KAC:1283383.1283494}
David Arthur and Sergei Vassilvitskii.
\newblock k-means++: the advantages of careful seeding.
\newblock In {\em Proceedings of the eighteenth annual ACM-SIAM symposium on
  Discrete algorithms}, SODA '07, pages 1027--1035, Philadelphia, PA, USA,
  2007. Society for Industrial and Applied Mathematics.

\bibitem{sparseCodingHandwriting2012}
Q.~Barthelemy, A.~Larue, A.~Mayoue, D.~Mercier, and J.I. Mars.
\newblock Shift and 2d rotation invariant sparse coding for multivariate
  signals.
\newblock {\em Signal Processing, IEEE Transactions on}, 60(4):1597--1611,
  2012.

\bibitem{DBLP:conf/sdm/BatistaWK11}
Gustavo E. A. P.~A. Batista, Xiaoyue Wang, and Eamonn~J. Keogh.
\newblock A complexity-invariant distance measure for time series.
\newblock In {\em SDM}, pages 699--710. SIAM / Omnipress, 2011.

\bibitem{cid2013}
Gustavo~E.A.P.A. Batista, Eamonn~J. Keogh, Oben~Moses Tataw, and Vinicius~M.A.
  Souza.
\newblock Cid: an efficient complexity-invariant distance for time series.
\newblock {\em Data Mining and Knowledge Discovery}, pages 1--36, 2013.

\bibitem{tsbf2013}
M.~Baydogan, G.~Runger, and E.~Tuv.
\newblock A bag-of-features framework to classify time series.
\newblock {\em Pattern Analysis and Machine Intelligence, IEEE Transactions
  on}, PP(99):1--1, 2013.

\bibitem{DBLP:conf/gfkl/BuzaS08}
Krisztian Buza and Lars Schmidt-Thieme.
\newblock Motif-based classification of time series with bayesian networks and
  svms.
\newblock In Andreas~Fink et~al., editor, {\em GfKl}, pages 105--114. Springer,
  2008.

\bibitem{Chen:2004:MLE:1316689.1316758}
Lei Chen and Raymond Ng.
\newblock On the marriage of lp-norms and edit distance.
\newblock In {\em Proceedings of the Thirtieth international conference on Very
  large data bases - Volume 30}, VLDB '04, pages 792--803. VLDB Endowment,
  2004.

\bibitem{Chen2007}
Yueguo Chen, M.A. Nascimento, Beng-Chin Ooi, and A.~Tung.
\newblock Spade: On shape-based pattern detection in streaming time series.
\newblock In {\em Data Engineering, 2007. ICDE 2007. IEEE 23rd International
  Conference on}, pages 786--795, 2007.

\bibitem{ICML2011Cuturi}
Marco Cuturi.
\newblock Fast global alignment kernels.
\newblock In Getoor et~al., editor, {\em Proceedings of the ICML 2011}, ICML
  2011, pages 929--936, New York, NY, USA, June 2011. ACM.

\bibitem{Demsar:2006:SCC:1248547.1248548}
Janez Dem\v{s}ar.
\newblock Statistical comparisons of classifiers over multiple data sets.
\newblock {\em J. Mach. Learn. Res.}, 7:1--30, December 2006.

\bibitem{ding2008}
Hui Ding, Goce Trajcevski, Peter Scheuermann, Xiaoyue Wang, and Eamonn~J.
  Keogh.
\newblock Querying and mining of time series data: experimental comparison of
  representations and distance measures.
\newblock {\em PVLDB}, 1(2):1542--1552, 2008.

\bibitem{DBLP:conf/aaai/GrabockaNS12}
Josif Grabocka, Alexandros Nanopoulos, and Lars Schmidt-Thieme.
\newblock Classification of sparse time series via supervised matrix
  factorization.
\newblock In J{\"o}rg Hoffmann and Bart Selman, editors, {\em AAAI}. AAAI
  Press, 2012.

\bibitem{DBLP:conf/pkdd/GrabockaNS12}
Josif Grabocka, Alexandros Nanopoulos, and Lars Schmidt-Thieme.
\newblock Invariant time-series classification.
\newblock In Peter~A. Flach, Tijl~De Bie, and Nello Cristianini, editors, {\em
  ECML/PKDD (2)}, volume 7524 of {\em Lecture Notes in Computer Science}, pages
  725--740. Springer, 2012.

\bibitem{Gudmundsson2008}
Steinn Gudmundsson, Thomas~Philip Runarsson, and Sven Sigurdsson.
\newblock Support vector machines and dynamic time warping for time series.
\newblock In {\em IJCNN}, pages 2772--2776. IEEE, 2008.

\bibitem{hills2013transform}
J.~Hills, J.~Lines, E.~Baranauskas, J.~Mapp, and A.~Bagnall.
\newblock Classification of time series by shapelet transformation.
\newblock {\em Data Mining and Knowledge Discovery, accepted subject to minor
  corrections}, 2013.

\bibitem{conf/interspeech/HuangYHLH12}
Po-Sen Huang, Jianchao Yang, Mark Hasegawa-Johnson, Feng Liang, and Thomas~S.
  Huang.
\newblock Pooling robust shift-invariant sparse representations of acoustic
  signals.
\newblock In {\em INTERSPEECH}. ISCA, 2012.

\bibitem{dimRedTimeSeries2001}
Eamonn Keogh, Kaushik Chakrabarti, Michael Pazzani, and Sharad Mehrotra.
\newblock Dimensionality reduction for fast similarity search in large time
  series databases.
\newblock {\em Knowledge and Information Systems}, 3(3):263--286, 2001.

\bibitem{Keogh2000}
Eamonn~J. Keogh and Michael~J. Pazzani.
\newblock Scaling up dynamic time warping for datamining applications.
\newblock In {\em KDD}, pages 285--289, 2000.

\bibitem{kuksaSSSK2010}
P.P. Kuksa and V.~Pavlovic.
\newblock Spatial representation for efficient sequence classification.
\newblock In {\em Pattern Recognition (ICPR), 2010 20th International
  Conference on}, pages 3320--3323, 2010.

\bibitem{LEW99}
Michael~S. Lewicki and Terrence~J. Sejnowski.
\newblock {Coding time-varying signals using sparse, shift-invariant
  representations}.
\newblock In {\em Proceedings of NIPS}, pages 730--736, Cambridge, MA, USA,
  1999. MIT Press.

\bibitem{Lin:2007:ESN:1285960.1285965}
Jessica Lin, Eamonn Keogh, Li~Wei, and Stefano Lonardi.
\newblock Experiencing sax: a novel symbolic representation of time series.
\newblock {\em Data Min. Knowl. Discov.}, 15(2):107--144, October 2007.

\bibitem{DBLP:journals/jiis/0001KL12}
Jessica Lin, Rohan Khade, and Yuan Li.
\newblock Rotation-invariant similarity in time series using bag-of-patterns
  representation.
\newblock {\em J. Intell. Inf. Syst.}, 39(2):287--315, 2012.

\bibitem{Lin:2009:FSS:1561638.1561679}
Jessica Lin and Yuan Li.
\newblock Finding structural similarity in time series data using
  bag-of-patterns representation.
\newblock In {\em Proceedings of the 21st International Conference on
  Scientific and Statistical Database Management}, SSDBM 2009, pages 461--477,
  Berlin, Heidelberg, 2009. Springer-Verlag.

\bibitem{DBLP:conf/kdd/MueenKY11}
Abdullah Mueen, Eamonn~J. Keogh, and Neal Young.
\newblock Logical-shapelets: an expressive primitive for time series
  classification.
\newblock In Chid Apt{\'e}, Joydeep Ghosh, and Padhraic Smyth, editors, {\em
  KDD}, pages 1154--1162. ACM, 2011.

\bibitem{Platt:1999:FTS:299094.299105}
John~C. Platt.
\newblock Advances in kernel methods.
\newblock chapter Fast training of support vector machines using sequential
  minimal optimization, pages 185--208. MIT Press, Cambridge, MA, USA, 1999.

\bibitem{rakthanmanon2013fast}
T.~Rakthanmanon and E.~Keogh.
\newblock Fast shapelets: A scalable algorithm for discovering time series
  shapelets.
\newblock {\em Proceedings of the 13th {SIAM} International Conference on Data
  Mining}, 2013.

\bibitem{Rakthanmanon2012}
Thanawin Rakthanmanon, Bilson Campana, Abdullah Mueen, Gustavo Batista, Brandon
  Westover, Qiang Zhu, Jesin Zakaria, and Eamonn Keogh.
\newblock Searching and mining trillions of time series subsequences under
  dynamic time warping.
\newblock In {\em Proceedings of the 18th ACM SIGKDD}, KDD 2012, pages
  262--270, New York, NY, USA, 2012. ACM.

\bibitem{Vlachos2002}
M.~Vlachos, G.~Kollios, and D.~Gunopulos.
\newblock Discovering similar multidimensional trajectories.
\newblock In {\em Data Engineering, 2002. Proceedings. 18th International
  Conference on}, pages 673--684, 2002.

\bibitem{Wang:2012:THT:2339530.2339605}
Fei Wang, Noah Lee, Jianying Hu, Jimeng Sun, and Shahram Ebadollahi.
\newblock Towards heterogeneous temporal clinical event pattern discovery: a
  convolutional approach.
\newblock In {\em Proceedings of ACM SIGKDD}, KDD '12, pages 453--461, New
  York, NY, USA, 2012. ACM.

\bibitem{Wang2013634}
Jin Wang, Ping Liu, Mary~F.H. She, Saeid Nahavandi, and Abbas Kouzani.
\newblock Bag-of-words representation for biomedical time series
  classification.
\newblock {\em Biomedical Signal Processing and Control}, 8(6):634 -- 644,
  2013.

\bibitem{surveyRepresentations2013}
Xiaoyue Wang, Abdullah Mueen, Hui Ding, Goce Trajcevski, Peter Scheuermann, and
  Eamonn Keogh.
\newblock Experimental comparison of representation methods and distance
  measures for time series data.
\newblock {\em Data Mining and Knowledge Discovery}, 26(2):275--309, 2013.

\bibitem{Zhang2010}
Dongyu Zhang, Wangmeng Zuo, David Zhang, and Hongzhi Zhang.
\newblock Time series classification using support vector machine with gaussian
  elastic metric kernel.
\newblock In {\em ICPR}, pages 29--32. IEEE, 2010.

\end{thebibliography}
\bibliographystyle{plain}

\end{document}